\documentclass{article}

\usepackage{arxiv}

\usepackage[utf8]{inputenc} % allow utf-8 input
\usepackage[T1]{fontenc}    % use 8-bit T1 fonts
\usepackage{hyperref}       % hyperlinks
\usepackage{url}            % simple URL typesetting
\usepackage{booktabs}       % professional-quality tables
\usepackage{amsfonts}       % blackboard math symbols
\usepackage{nicefrac}       % compact symbols for 1/2, etc.
\usepackage{microtype}      % microtypography
\usepackage{lipsum}		% Can be removed after putting your text content
\usepackage{graphicx}
\usepackage{natbib}
\usepackage{doi}

\usepackage{graphicx}
\usepackage{amsmath}
\usepackage{colortbl}
\usepackage{array}
\usepackage{booktabs}
\usepackage{colortbl}
\usepackage{xcolor}
\usepackage{multirow}
\usepackage{amssymb}
\usepackage{subcaption}
\usepackage{tabularray}
\usepackage{colortbl}
\usepackage{tabularx}
\usepackage{tikz-dependency}
\usepackage{subcaption}
\usepackage{float}

\usepackage{caption}
\usepackage{adjustbox}

\definecolor{aspectColor}{rgb}{0.6,0.8,1}        % Soft blue for aspect term
\definecolor{positiveGreen}{rgb}{0.7,1,0.7}       % Soft green for aspect class
\definecolor{opinionColor}{rgb}{1,0.8,0.6}      % Soft orange for opinion term
\definecolor{aspectClass}{rgb}{0.7,1,0.9}     % Another shade of soft green
\definecolor{negativeRed}{rgb}{1,0.7,0.7}       % Soft red

\title{Exploiting Adaptive Contextual Masking for Aspect-Based Sentiment Analysis}

\author{ \href{https://orcid.org/0000−0001−9404−1556}{\includegraphics[scale=0.06]{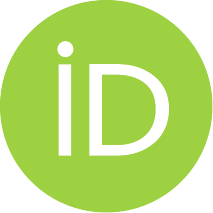}\hspace{1mm}S M Rafiuddin}\thanks{Corresponding Author} \\
	Department of Computer Science\\
	Oklahoma State University\\
	Stillwater, Oklahoma, USA \\
	\texttt{srafiud@okstate.edu} \\
	\And
	\href{https://orcid.org/0000-0001-6201-3729}{\includegraphics[scale=0.06]{orcid.pdf}\hspace{1mm}Mohammed Rakib} \\
	Department of Computer Science\\
	Oklahoma State University\\
	Stillwater, Oklahoma, USA  \\
	\texttt{mohammed.rakib@okstate.edu} \\
    \And
	\href{https://orcid.org/0000-0003-0353-0894}{\includegraphics[scale=0.06]{orcid.pdf}\hspace{1mm}Sadia Kamal} \\
	Department of Computer Science\\
	Oklahoma State University\\
	Stillwater, Oklahoma, USA  \\
	\texttt{sadia.kamal@okstate.edu} \\
 \And
	\href{https://orcid.org/0000-0002-7135-4602}{\includegraphics[scale=0.06]{orcid.pdf}\hspace{1mm}Arunkumar Bagavathi} \\
	Department of Computer Science\\
	Oklahoma State University\\
	Stillwater, Oklahoma, USA  \\
	\texttt{abagava@okstate.edu} \\
}

\renewcommand{\headeright}{PAKDD 2024}
\renewcommand{\undertitle}{Accecpted at 28\textsuperscript{th} Pacific-Asia Conference on Knowledge Discovery and Data Mining}
\renewcommand{\shorttitle}{Exploiting Adaptive Contextual Masking for ABSA}

\hypersetup{
pdftitle={A template for the arxiv style},
pdfsubject={q-bio.NC, q-bio.QM},
pdfauthor={David S.~Hippocampus, Elias D.~Striatum},
pdfkeywords={First keyword, Second keyword, More},
}

\begin{document}
\maketitle

\begin{abstract}
Aspect-Based Sentiment Analysis (ABSA) is a fine-grained linguistics problem that entails the extraction of multifaceted aspects, opinions, and sentiments from the given text. Both \emph{standalone} and \emph{compound} ABSA tasks have been extensively used in the literature to examine the nuanced information present in online reviews and social media posts. Current ABSA methods often rely on static hyperparameters for attention-masking mechanisms, which can struggle with context adaptation and may overlook the unique relevance of words in varied situations. This leads to challenges in accurately analyzing complex sentences containing multiple aspects with differing sentiments. In this work, we present adaptive masking methods that remove irrelevant tokens based on context to assist in Aspect Term Extraction and Aspect Sentiment Classification subtasks of ABSA. We show with our experiments that the proposed methods outperform the baseline methods in terms of accuracy and F1 scores on four benchmark online review datasets. Further, we show that the proposed methods can be extended with multiple adaptations and demonstrate a qualitative analysis of the proposed approach using sample text for aspect term extraction.\footnote{For code and dataset, please contact: \textit{\textbf{srafiud@okstate.edu}}}
\end{abstract}

% keywords can be removed


\begin{figure}[H]
  \centering
 % Set the image width to the column width
  \includegraphics[width=0.9\textwidth]{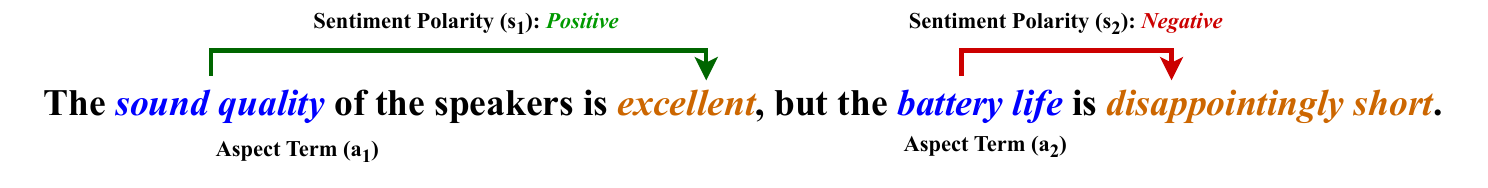}
  \caption{Simple ABSA for an online review. The review consists of multiple \textcolor{blue}{\textit{\textbf{aspect terms}}} and each aspect of the review contains its own \textcolor{orange}{\textit{\textbf{sentiment polarity}}}.}
  \label{fig:ABSA}
\end{figure}

\section{Introduction}
Aspect-Based Sentiment Analysis (ABSA) tasks are gaining traction in various online domains like customer reviews and social media monitoring~\cite{zhang2022survey}. Traditional sentiment analysis tasks consider only one sentiment polarity label, \emph{positive}, \emph{neutral}, or \emph{negative}, for the given text. However, the real-world text may consist of multiple sentiment polarities assigned to multifaceted aspects and opinions. A simple example of the ABSA task on a restaurant review is depicted in Figure~\ref{fig:ABSA}. ABSA tasks usually involve \emph{four} components: \emph{aspect category}, \emph{aspect terms}, \emph{opinion terms}, and \emph{sentiment polarity}. Several works are available in the literature to study ABSA tasks in both \emph{standalone}~\cite{phan-ogunbona-2020-modelling} and \emph{compound}~\cite{lin2022aspect} fashion. \emph{Standalone} approaches aim to extract only one component, whereas the \emph{compound} approaches jointly extract more than one component, for example extracting aspect term and sentiment. In this paper, we focus on two standalone ABSA tasks of \emph{aspect term extraction} (ATE) and \emph{aspect sentiment classification} (ASC). ASC is a supervised task that predicts the sentiment polarity of the given text in the context of a given topic or aspect. As given in Figure~\ref{fig:ABSA}, the text data can contain multiple aspects and each aspect can have a distinct sentiment polarity. Similarly, the ATE is a supervised task to extract the start and end positions of each aspect available in the text. It is common that there are multiple aspect terms and each aspect term can span across multiple words of the given text, similar to the example depicted in Figure~\ref{fig:ABSA}.

The introduction of attention~\cite{vaswani2017attention} and large language models increased the scope of sentiment analysis problems by understanding the precise context of the text. Attention and LLMs aid ABSA tasks in capturing \emph{local} context of words and \emph{global} contextual features of the entire text respectively. The self-attention strategy aims to extract useful information from text with respect to words present in the text itself. Such techniques are crucial for any ABSA tasks to map aspect categories and aspect terms with opinion terms and sentiment. One of the popular approaches with ABSA is to filter out noisy terms that do not cover contextual details for other relevant terms of the given text. Such strategies also utilize attention weights that are optimized to \emph{masked} words that are not useful or out of context to the given aspect. In the literature, the normalized attention weights have been used in two different forms. (1) \emph{Weights Threshold:} Threshold-based approaches set a user-defined threshold or take the maximum weight as the threshold~\cite{feng2022unrestricted}. (2) \emph{Distance Based:} These approaches sort attention weights and define a window size around the aspect terms ~\cite{phan-ogunbona-2020-modelling}. If the word attention weights do not satisfy the threshold or distance condition, they are considered noisy terms and \emph{masked} with zero-vectors for any downstream ABSA task. Recently, dynamic approaches to extract correlations between the local context and aspects of the text are gaining traction in ABSA problems~\cite{zeng2019lcf,phan-ogunbona-2020-modelling,feng2022unrestricted}. However, these approaches are introduced with an assumption of a user-assigned threshold to decide the tokens to mask. In addition to the local context, the pre-trained LLMs give better contextual features of the text to cover global representations for ABSA tasks. It is a common practice to use both attention and LLM representations in tandem to perform both \emph{standalone}~\cite{lin2023adaptive} and \emph{compund} ABSA tasks. In this work, we present three key contributions to overcome the challenges of current dynamic approaches in standalone ABSA tasks:

\begin{enumerate}
    \item \textbf{Adaptive Contextual Threshold Masking (ACTM)}: We introduce a novel masking strategy that adjusts mask thresholds adaptively to determine the mask ratio of text tokens based on their context and enhance granularity in standalone ABSA tasks.

    \item \textbf{Adaptive Masking for ABSA}: In addition to the proposed ACTM strategy, we tailored two existing distance-based adaptive attention masking techniques exclusively for standalone ABSA tasks.

    \item \textbf{Experimental Validation}: We demonstrate with extensive experiments that the adaptive masking approaches outperform baseline ABSA methods across multiple \emph{SemEval} benchmark datasets for two ABSA subtasks.
\end{enumerate}

\section{Related Work}

Recent advancements in Aspect-Based Sentiment Analysis (ABSA) include joint learning approaches by Mao et al.~\cite{mao2021joint} using shared pre-trained models and Xu et al.~\cite{xu-etal-2021-learning} employing sequence-to-sequence models. Additionally, Yan et al.~\cite{yan-etal-2021-unified} and Zhang et al.~\cite{zhang2021towards} have proposed unified generative frameworks, integrating Pre-trained Language Models (PLMs) and treating ABSA tasks as distinct text generation challenges.

Key subtasks in ABSA, such as aspect term extraction and sentiment polarity determination, are highlighted in Chen and Qian~\cite{chen-qian-2020-enhancing}. These tasks increasingly rely on aspect-based syntactic information (POS-tags), as evidenced in works like~\cite{ijcai2021p545,10.1145/3511808.3557452}. Phan~\cite{phan-ogunbona-2020-modelling} emphasizes the importance of context in ABSA by combining syntactical features with contextualized embeddings. Span-level models for aspect sentiment extraction are explored in Chen et al.~\cite{chen-etal-2022-span}, while GCN-based models for extracting syntactic information from dependency trees are discussed in Zhang et al.~\cite{zhang-etal-2022-ssegcn}. Additionally, Tian et al.~\cite{tian2021aspect} use GNNs over dependency trees to enhance understanding of syntactic features.

Recent ABSA research emphasizes attention-based neural networks~\cite{feng2022unrestricted,liu2015fine}, with key advancements like Feng et al.'s masked attention method (AM-WORD-BERT) for term-focused performance enhancement~\cite{feng2022unrestricted}. Lin et al.'s AMA-GLCF model~\cite{lin2023adaptive} utilizes a masked attention mechanism with Context Dynamic Mask (CDM) and Context Dynamic Weight (CDW) for prioritizing aspect-relevant text in global and local contexts. Additionally, LCFS uses CDW and CDM for improved aspect extraction and sentiment classification by leveraging attention-based masking~\cite{phan-ogunbona-2020-modelling}. In these models, they have used a static threshold for masking, but we proposed an adaptive masking based on the context. Adaptive threshold masking enhances model accuracy by adjusting to the varying relevance of data features in different contexts. It improves overall model performance by ensuring focus on the most relevant aspects of the input.

\section{Methodology}

\begin{figure}[htbp]
\small
  \centering
  \begin{subfigure}[b]{0.45\textwidth}
    \centering
    \includegraphics[width=\textwidth]{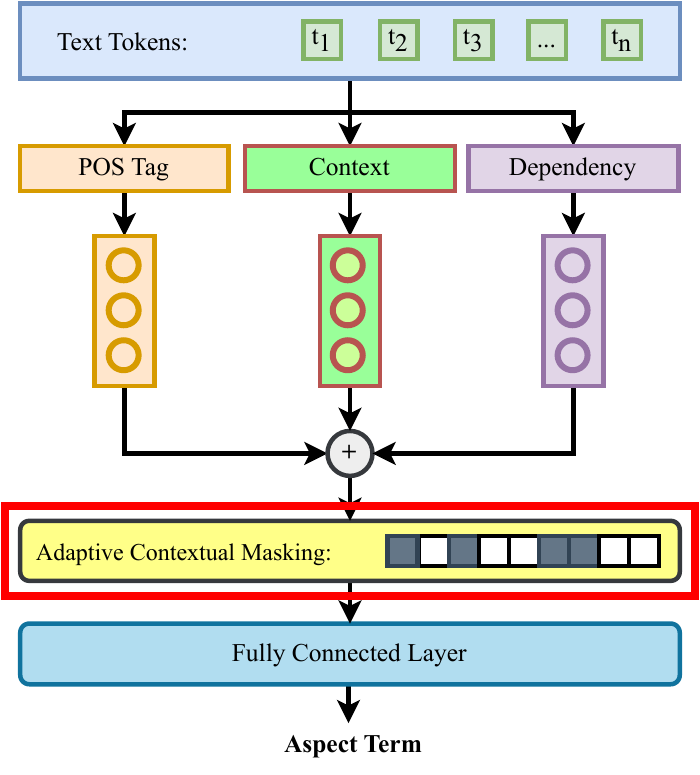}
    \caption{Aspect Term Extraction}
    \label{fig:aspect_term_extraction}
  \end{subfigure}%
  \hfill
  \begin{subfigure}[b]{0.45\textwidth}
    \centering
    \includegraphics[width=\textwidth]{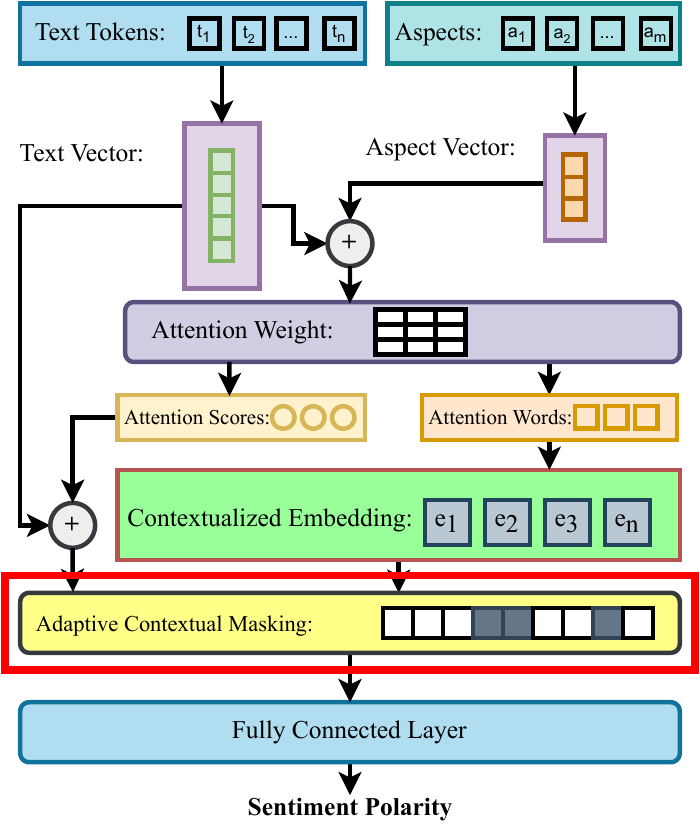}
    \caption{Aspect Sentiment Classification}
    \label{fig:aspect_sentiment_classification}
  \end{subfigure}
  \caption{Proposed Structure of (a) Aspect Term Extraction, and (b) Aspect Sentiment Classification. The \textcolor{red}{\textbf{red box}} highlights the models introduced in this paper.}
  \label{fig:sample_svg}
\end{figure}

\subsection{Problem Formulation and Motivation}

%Given a text instance \( T \) with \emph{n} tokens $\{T_i\}_{i=1}^n$ \( \{T_1, T_2, \ldots, T_n\} \), where each \( T_i \in T \), and the set of aspect terms given as \( A_T \), with sentiment polarity \( P_A = \{\text{pos}, \text{neg}, \text{neutral}\} \), define the following tasks:

% are considered as the text instance $T$'s aspect terms (\emph{perspectives}) and
Given a text $T = \{t_i\}_{i=1}^n = \{t_1, t_2, \ldots, t_n\}$ of \emph{n} tokens with $k$ aspects $\{\mathcal{A}_j\}_{j=1}^k$ where each aspect spans across multiple tokens in $T$ and each aspect can map to its own sentiment polarity $\mathcal{A}_j \rightarrow \mathcal{P}_j \in \{pos., neut., neg.\}$. We define the aspect term extraction task as a supervised approach $f_{\text{ATE}}: \mathcal{M}(\mathbb{E}(T), t_i) \rightarrow \hat{\mathcal{P}}_{t_i} \in \{Begin, In, Out\}$ and the aspect sentiment classification task as $f_{\text{ASC}}: \mathcal{M}(\mathbb{E}(T), \mathbb{E}(\mathcal{A}_j)) \rightarrow \hat{\mathcal{P}}_j$ where $\mathbb{E}(T) \in \mathbb{R}^{n \times d}$ is token representations, $\mathbb{E}(\mathcal{A}_j) \in \mathbb{R}^{m \times d} | m<<<n$ is aspect representations, $\hat{\mathcal{P}}_{w_i}$ is the predicted position of $w_i$, and $\hat{\mathcal{P}}_j$ is the predicted sentiment polarity. In this work, we contribute with the adaptive contextual masking $\mathcal{M}: \mathbb{E}(T) \rightarrow \mathbb{E}^\prime(T)$ for the above given two ABSA subtasks, where $\mathbb{E}^\prime = \{[MASK], e_2,[MASK], \ldots, e_n\}$ is a masked representations of $\mathbb{E}$. Our motivation for this work is to design an adaptive masking $\mathcal{M}$ that adjusts token masks to hide irrelevant terms in correspondence to the context of $T$, rather than using a hard threshold as given by dynamic masking approaches. 

\subsection{Standalone ABSA Tasks}
We briefly outline \emph{aspect term extraction} (ATE) and \emph{aspect-based sentiment classification} (ASC) tasks, providing context for our proposed masking methods. Figure~\ref{fig:sample_svg} shows the model architectures, inspired by established approaches~\cite{feng2022unrestricted,lin2023adaptive,phan-ogunbona-2020-modelling}.

% \subsubsection{Aspect Term Extraction (ATE)} As depicted in Figure~\ref{fig:aspect_term_extraction}, our aspect term extraction architecture takes the input data and predicts the \emph{start} and \emph{end} positions of aspect terms present in $T$. The proposed architecture utilizes \emph{three} types of features to learn input text representations. \emph{\textbf{First}}, we use POS tags that map the given text into a $d$-dimensional POS tag vector. POS tags in turn capture basic syntactic features of tokens which are instrumental in identifying aspect terms. \emph{\textbf{Second}}, we use contextual features to capture the context of the text data. \emph{\textbf{Third}}, we utilize complex syntactic features using relationships of tokens present in dependency graphs. We concatenate these three vectors to represent the feature vector of the text $T$. We extract aspect terms of $T$ using a fully connected output layer after masking irrelevant tokens using the proposed \emph{DTM} strategy (given in Section~\ref{sect:DTM}).

\subsubsection{Aspect Term Extraction (ATE)}
The ATE model, depicted in Figure~\ref{fig:aspect_term_extraction}, predicts aspect terms in text $T$. The input $T$ is formatted as \texttt{[CLS] + T + [SEP]}, where \texttt{[CLS]} and \texttt{[SEP]} define the sentence context and endpoint. We utilize POS tag representations ($p_i$), contextual token features ($w_i$), and dependency graph-based syntactic features ($D_i$) of tokens $\{t_i\}_{i=1}^n \in T$. We concatenate these grammatic, contextual and syntactic features to represent each token representation as $\mathbf{e}_i = \mathbf{w}_i \oplus \mathbf{p}_i \oplus \mathbf{D}_i$. We then process $\mathbb{E}(T)=\{e_1, e_2, \ldots e_n\}$ through the adaptive contextual masking(Sections~\ref{sect:DTM}, \ref{sect:adaptive}, \ref{sect:amom}) and a fully connected layer to perform the supervised task $f_{ATE}$. 

\subsubsection{Aspect-Based Sentiment Classification (ASC)} The ASC model, depicted in Figure~\ref{fig:aspect_sentiment_classification}, predicts sentiment polarities of each aspect $\mathcal{A}_j \in T$. The ASC model analyze sentiment of $T$ towards the given specific aspect $\mathcal{A}_j$ by employing the input \texttt{[CLS]+$T$+[SEP]+$\mathcal{A}_j$+[SEP]}, where \texttt{[CLS]} captures overall context. We apply a transformer model, like BERT~\cite{kenton2019bert}, to extract contextual representations of $T$ and $\mathcal{A}_j$. These features are fed into the attention mechanism, which calculates the attention weights between the aspect vectors and the text vectors. The attention mechanism assigns higher weights to the words or tokens that are more relevant to the aspect of focus based on context, allowing the model to focus on the most important information for sentiment analysis. We mask the tokens in $\mathbb{E}(T)$ by measuring its relevance with $\mathbb{E}(\mathcal{A}_j)$ using the adaptive contextual masking(Sections~\ref{sect:DTM}, \ref{sect:adaptive}, \ref{sect:amom}) and perform the supervised task $f_{ASC}$.

\subsection{Adaptive Contextual Threshold Masking (ACTM)}
\label{sect:DTM}

\begin{figure}[htbp]
  \centering
 % Set the image width to the column width
  \includegraphics[scale=0.60]{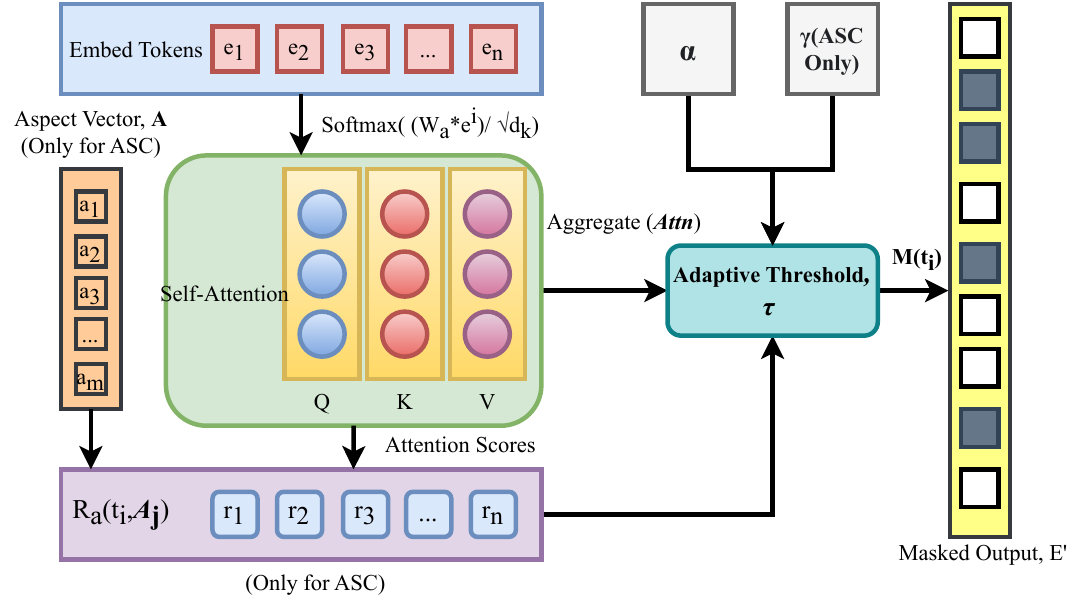}
  \caption{Adaptive Contextual Threshold Masking strategy to adjust the masking threshold $\tau$ using the aggregated attention scores of token representations. $\alpha$ and $\gamma$ are the learnable parameters, where $\gamma$ is used in
 the ASC task only for aspect relevance.}
  \label{fig:svgImage}
\end{figure}

We propose the ACTM strategy, given in Figure~\ref{fig:svgImage}, to dynamically adjust the attention span and update mask threshold to prioritize the sentiment-bearing tokens related to aspect terms. First, we capture token significance in $T$ by computing attention scores $\textit{\textbf{Attn}}=\{attn_1, attn_2, \ldots attn_n\}$ of token representations $\{e_i\}_{i=1}^n \in \mathbb{E}(T)$ using the self attention mechanism, as given in Equation~\ref{eq:self_attn}. 
%For a given text \( T \), we compute the contextual embeddings \( \mathbf{E} = [e_1, e_2, \ldots, e_n] \) for the tokens \( T = [t_1, t_2, \ldots, t_n] \). The attention mechanism then computes scores \( \mathbf{A} = [a_1, a_2, \ldots, a_n] \) with:

\begin{equation}
    attn_i = \text{softmax}\left(\frac{W_a e_i}{\sqrt{d_k}}\right)
    \label{eq:self_attn}
\end{equation}

where \( W_a \) is the weight matrix for the attention layer and \( d_k \) is the dimensionality of the key vectors in the attention mechanism. For the ATE task, we incorporate the adaptive threshold $\tau$ on each token $t_i$ as $\tau(t_i) = \alpha \cdot \text{\textbf{Aggregate}}\left(\textit{\textbf{Attn}}\right)$. Similarly, for the ASC task, we employ the adaptive threshold on tokens as $\tau(t_i) = \alpha \cdot \text{\textbf{Aggregate}}\left(\textit{\textbf{Attn}}\right) + \gamma \cdot R_a(t_i, \mathcal{A}_j)$, where \( R_a(t_i, \mathcal{A}_j) \) is the aspect relevance function that measures the relevance of token \( t_i \) to the given aspect $\mathcal{A}_j$ using the attention scores as given in Equation~\ref{eq:relevance}. The \( \alpha \) and \( \gamma \) are learnable parameters that adjust the influence of the attention scores and aspect relevance, respectively. \textbf{Aggregate} is a pooling function that acts as a metric for getting the most relevant terms based on context. 

%To adapt the attention scores to aspect terms, we used an aspect relevance function \( R_a(t_i, A) \) (Only for ASC task), which measures the relevance of token \( t_i \) to the set of aspect terms \( A \):

\begin{equation}
    R_a(t_i, \mathcal{A}_j) = \frac{\exp\left(\beta \cdot \text{\textbf{sim}}(a_i, \mathbb{E}(\mathcal{A}_j)\right)}{\sum_{k=1}^{n}\exp\left(\beta \cdot \text{\textbf{sim}}(a_k, \mathbb{E}(\mathcal{A}_k))\right)}
    \label{eq:relevance}
\end{equation}

where $a_i$ is weighted contextual token vector, $\mathbb{E}(\mathcal{A}_j)$ represents the contextual aspect vector, \( \text{\textbf{sim}} \) is a \textbf{\textit{cosine similarity}} function, and \( \beta \) is a scaling factor. The irrelevant tokens are masked for the sentiment analysis by matching their corresponding attention scores with adaptive threshold $\tau$ as given in Equation~\ref{eq:mask}.

%The dynamic threshold \( \tau \) incorporates aspect relevance and is computed as:

%For ATE:

%\begin{equation}
%    \tau(t_i) = \alpha \cdot \text{\textbf{Aggregate}}\left(\mathbf{A}\right) 
%\end{equation}

%For ASC:
%\begin{equation}
%    \tau(t_i) = \alpha \cdot \text{\textbf{Aggregate}}\left(\mathbf{A}\right) + \gamma \cdot R_a(t_i, A)
%\end{equation}

%where \( \alpha \) and \( \gamma \) are learnable parameters that adjust the influence of the attention scores and aspect relevance, respectively. 
%\textbf{Aggregate} would be any functions like- Mean, Standard Deviation (SD), etc. Tokens are masked or retained for the sentiment analysis based on their thresholded attention scores:

\begin{equation}
    \mathcal{M}(t_i) = 
    \begin{cases}
      attn_i, & \text{if}\ attn_i \geq \tau(t_i) \\
      0, & \text{otherwise}
    \end{cases}
    \label{eq:mask}
\end{equation}

This mask $\mathcal{M}$ results in a filtered context $\mathbb{E}^\prime$ that emphasizes aspect-related sentiments. The iterative process allows the threshold \( \tau \) to adapt dynamically, providing a focused analysis of the interplay between aspects and sentiments in varying contexts. By employing contextual adaptive masking, ABSA models can effectively concentrate on the most informative tokens that influence the sentiment towards their aspect for precision ABSA tasks.

\subsection{Adaptive Attention Masking (AAM)}

\label{sect:adaptive}

Sukhbaatar \textit{et al.}~\cite{sukhbaatar2019adaptive} introduced an adaptive self-attention mechanism with a soft masking function and a masking ratio \( M \) in transformer models. This mechanism dynamically adjusts attention spans to emphasize relevant terms for any downstream tasks. The soft masking function in self-attention layers, \( m_z(x) = \min \left[ \max \left[ \frac{1}{R} (R + z - x), 0 \right], 1 \right] \), where \( R \) is a flexibility hyper-parameter and \( z \) is a learnable parameter, helps in this dynamic adjustment. \( x \) is the distance from a given position in a sequence of tokens to the position being focused on by the attention mechanism. The adaptive masking ratio \( M \), crucial for determining the span of attention, is calculated as \( \mathcal{M} = \frac{\sum m_z(x)}{n} \). Consequently, the attention weights are given by the equation \(\text{Attention Weights} = \text{softmax} \left( \frac{m_z(x) \cdot \mathcal{M}}{\sqrt{d_k}} \right)\), integrating \( \mathcal{M} \) and using scaled softmax within the span boundaries \( l \) and \( r \). Given a current position \( p \) in the sequence and a learned attention span \( z \), the left boundary \( l \) is calculated as \( l = p - z \), and the right boundary \( r \) is calculated as \( r = p + z \), defining the span of attention around position \( p \). This adaptive attention approach, by incorporating the masking ratio, allows the model to capture extended dependencies, which is vital for effective ATE and ASC tasks.

\subsection{Adaptive Mask Over Masking (AMOM)}

\label{sect:amom}

Xiao \textit{et al.}~\cite{xiao2023amom} introduced Adaptive Masking Over Masking (AMOM), which is adapted to enhance both Aspect Term Extraction (ATE) and Aspect Sentiment Classification (ASC) in conditional masked language models (CMLM). In our tasks of ABSA, AMOM generates masked sequences \( Y_{\text{mask}}^{\text{ATE}} \) and \( Y_{\text{mask}}^{\text{ASC}} \) using input, aspect (Only for ASC), and sentiment labels. It evaluates prediction correctness for ATE and ASC against the ground truth \( Y \), calculating correctness ratios \( R_{\text{ATE}} \) and \( R_{\text{ASC}} \). These ratios inform adaptive masking ratios \( \mu_{\text{ATE}} = f(R_{\text{ATE}}) \) and \( \mu_{\text{ASC}} = f(R_{\text{ASC}}) \), with \( N_{\text{mask}}^{\text{ATE}} = \mu_{\text{ATE}} \cdot |Y| \) and \( N_{\text{mask}}^{\text{ASC}} = \mu_{\text{ASC}} \cdot |Y| \) indicating the number of tokens to mask and regenerate for each task. The regenerating processes \( Y'_{\text{ATE}} = G(Y_{\text{mask}}^{\text{ATE}}, X) \) and \( Y'_{\text{ASC}} = G(Y_{\text{mask}}^{\text{ASC}}, X) \) adapt to the ABSA context, with \( M(X, R_{\text{ATE}}) \) and \( M(X, R_{\text{ASC}}) \) dynamically adjusting the decoder’s masking strategy for both ATE and ASC, effectively addressing nuanced expressions related to specific aspects and sentiments in text. We highlight that both AAM and AMOM have never been explored for any ABSA tasks and we combine these strategies with both ATE and ASC tasks in this work.

\subsection{Training Procedure for ATE and ASC}
\label{sect:training}

We train the ATE task by minimizing the categorical cross-entropy loss employed with the BIO (Begin, Inside, Outside) tagging scheme as given in Equation~\ref{eq:ATE_loss}.
\begin{equation}
L = -\sum_{i=1}^{N} \sum_{c \in \{B, I, O\}} y_{i,c} \log(\hat{y}_{i,c})
\label{eq:ATE_loss}
\end{equation}
where \( N \) is the number of words, \( y_{i,c} \) is the trinary indicator of the class label \( c \) for the word \( i \), and \( \hat{y}_{i,c} \) is the predicted probability. We train the ASC task using the multi-class loss function with \( L2 \) regularization as given in Equation~\ref{eq:ASC_loss}.
%for word \( i \) in class \( c \). 
%The goal is to minimize this loss, focusing on BIO tag prediction and model optimization.

%For Aspect Sentiment Classification (ASC), a multi-class loss function with \( L2 \) regularization is used:
\begin{equation}
L(\mathbf{Y}, \mathbf{\hat{Y}}, \Theta) = -\frac{1}{n} \sum_{i=1}^n \sum_{c=1}^C y_{ic} \log(\hat{y}_{ic}) + \frac{\lambda}{2} \|\Theta\|_2^2
\label{eq:ASC_loss}
\end{equation}
where \( \mathbf{Y} \) and \( \mathbf{\hat{Y}} \) are  true class and predicted class probabilities respectively, \( C \) is the class count, \( y_{ic} \) and \( \hat{y}_{ic} \) denote the true and predicted class probabilities for instance \( i \), respectively, and \( \|\Theta\|_2^2 \) is the \( L2 \) norm of the model parameters \( \Theta \).

%, \( \mathbf{\hat{Y}} \) is the matrix of predicted class probabilities, \( n \) is the total number of instances

%This approach balances fit and regularization to prevent overfitting.

\section{Experiments and Results}
% \subsection{Datasets}
% \begin{table*}
    
% \centering
% \caption{Data statistics} \label{table:dataset} 
% \begin{tabular}{|p{1.5cm}p{1.5cm} |p{1.5cm} | p{1.5cm}  p{1.5cm} p{1.5cm}|}
%  \hline
% \multicolumn{2}{|c|}{\multirow{2}{*}{Dataset}}& \multicolumn{1}{c}{\#Sent.}&\multicolumn{3}{c|}{\#Aspects} \\
% \cline{3-6}
% \multicolumn{1}{|l}{} & & & Positive & Negative & Neutral\\
%  \hline
%  14lap& Train & 1124 & 994& 870 & 464\\
%   & Test& 332& 341 & 128 &464\\
%  \hline
% 14res & Train & 1574&2164& 807 & 637\\ 
% & Test& 493& 728 & 196 & 196\\
% \hline
% 15res & Train& 721& 1777& 334& 81\\
%  & Test& 318 & 703& 192& 46\\
% \hline
% 16res & Train & 1052 & 2451 & 532 & 125\\
% & Test & 319 & 685 & 93 & 63\\
% \hline
% \end{tabular}
% \end{table*}

% \subsection{Datasets}

% \begin{table*}
% \centering
% \caption{Data statistics} \label{table:dataset} 
% \begin{tabular}{|c|c|c|c|c|c|}
%  \hline
% \multirow{2}{*}{Dataset} & \multirow{2}{*}{} & \#Sent. & \multicolumn{3}{c|}{\#Aspects} \\
% \cline{4-6}
%  & & & Pos. & Neg. & Neu. \\
%  \hline
% 14lap & Train & 1124 & 994 & 870 & 464 \\
%  & Test & 332 & 341 & 128 & 464 \\
%  \hline
% 14res & Train & 1574 & 2164 & 807 & 637 \\
%  & Test & 493 & 728 & 196 & 196 \\
%  \hline
% 15res & Train & 721 & 1777 & 334 & 81 \\
%  & Test & 318 & 703 & 192 & 46 \\
%  \hline
% 16res & Train & 1052 & 2451 & 532 & 125 \\
%  & Test & 319 & 685 & 93 & 63 \\
%  \hline
% \end{tabular}
% \end{table*}

%\subsection{Datasets}

\textbf{\textit{Datasets}:} We experiment with benchmark ABSA datasets from Semeval 2014\footnote{https://alt.qcri.org/semeval2014/task4/}, 2015\footnote{https://alt.qcri.org/semeval2015/task12/}, and 2016\footnote{https://alt.qcri.org/semeval2016/task5/}. Table~\ref{table:dataset} details the counts of reviews and the total number of positive, neutral, and negative aspects present in the overall training and test datasets used in our experiments. These training and test samples are pre-defined in SemEval and the same setup is utilized in all our baseline models.

%\subsection{Experiment Setup}
\noindent \textbf{\textit{Experiment Setup}:} 
We have leveraged BERT (``bert-base-cased'') model as our contextual feature extractor for all of the proposed approaches~\cite{kenton2019bert}. Layer normalization is set to \(1 \times 10^{-12}\) and a dropout of 0.1 is applied to the attention probabilities and hidden layers. For all experiments, we have trained for 50 epochs with a batch size of 32, a learning rate of \(2 \times 10^{-5}\), and L2 regularization of 0.01. Experiments were conducted in a lab server with 3$\times$NVIDIA A10 GPUs. We use \emph{Mean} for the \textbf{Aggregate} operation in the ACTM model.

\begin{table}[h]
\centering
\caption{Summary of laptop and restaurant review datasets given by SemEval.} \label{table:dataset} 
\renewcommand{\arraystretch}{1.0}
\begin{tabular}{|l|c|c|c|c|c|}
 \hline
\multirow{2}{*}{\textbf{Dataset}} & \multirow{2}{*}{\textbf{Split}}& \textbf{\#Reviews} & \multicolumn{3}{c|}{\textbf{\#Aspects}} \\
\cline{3-6}
 & & & \textbf{Positive} & \textbf{Negative} & \textbf{Neutral} \\
 \hline
Laptop14 & Train & 1124 & 994 & 870 & 464 \\
 & Test & 332 & 341 & 128 & 464 \\
 \hline
Restaurant14 & Train & 1574 & 2164 & 807 & 637 \\
 & Test & 493 & 728 & 196 & 196 \\
 \hline
Restaurant15 & Train & 721 & 1777 & 334 & 81 \\
 & Test & 318 & 703 & 192 & 46 \\
 \hline
Restaurant16 & Train & 1052 & 2451 & 532 & 125 \\
 & Test & 319 & 685 & 93 & 63 \\
 \hline
\end{tabular}
\end{table}

\noindent \textbf{\textit{Baseline Methods}:}\textbf{(1) ATE task: }
\textbf{BiLSTM}~\cite{liu2015fine} is an RNN used for NLP tasks, processing sequences bidirectionally. \textbf{DTBCSNN}~\cite{ye2017dependency} leverages sentence dependency trees and CNNs for aspect extraction. \textbf{BERT-AE}~\cite{kenton2019bert} uses BERT's pre-trained embeddings for aspect extraction. \textbf{IMN}~\cite{he2019interactive}, combines memory networks with aspect-context interactive attention. (\textbf{CSAE})~\cite{phan-ogunbona-2020-modelling} combines various components such as contextual, dependency-tree, and self-attention mechanisms. 
\textbf{(2) ASC task:}
Models such as \textbf{LCF-ASC-CDM}, \textbf{LCF-ASC-CDW}~\cite{zeng2019lcf}, \textbf{LCFS-ASC-CDW}, and \textbf{LCFS-ASC-CDM}~\cite{phan-ogunbona-2020-modelling} use Local Context Focus (LCF) with Context Dynamic Mask (CDM) and Weight (CDW) layers. Attention Mask variations like \textbf{AM Weight-BERT} and \textbf{AM Word-BERT} are applied in ABSA, targeting relevant parts~\cite{feng2022unrestricted}. The \textbf{Unified Generative} model~\cite{yan-etal-2021-unified} utilizes BART for multiple ABSA tasks. \textbf{MGGCN-BERT}~\cite{xiao2022multi} leverages BERT embeddings for ASC, while \textbf{AMA-GLCF}~\cite{lin2023adaptive} combines global and local text contexts using masked attention.

%%In our approach, along with these features, we applied Context-based adaptive masking to get the most relevant terms to extract aspects.

\begin{table}[h]
\centering
\caption{Adaptive contextual masking strategies against baseline models on ATE. "\textbf{-}" signifies no results available, and "\textbf{*}" denotes reproduced results.}
%, and bold highlights the best results for specific dataset/subtask.}
% For ACTM model, Gradient-based Mean is applied.
\label{tab:ate_datasets}
\footnotesize % Slightly smaller font size
\begin{tabular}{|l|c|c|c|c|}
\hline
\multirow{2}{*}{\textbf{Model}} & \multicolumn{1}{c|}{Laptop14} & \multicolumn{1}{c|}{Restaurant14} & \multicolumn{1}{c|}{Restaurant15} & \multicolumn{1}{c|}{Restaurant16} \\ \cline{2-5} 
                                & F1 & F1 & F1 & F1 \\ \hline

\textbf{BiLSTM~\cite{liu2015fine}} & 73.72 & 81.42 & - & - \\
\textbf{DTBCSNN~\cite{ye2017dependency}} & 75.66 & 83.97 & - & - \\
\textbf{BERT-AE~\cite{kenton2019bert}} & 73.92 & 82.56 & - & - \\
\textbf{IMN~\cite{he2019interactive}} & 77.96 & 83.33 & 70.04 & 78.07* \\
\textbf{CSAE~\cite{phan-ogunbona-2020-modelling}} & 77.65 & \textbf{86.65} & 76.84* & 80.63* \\
% \textbf{Unified Generative~\cite{yan-etal-2021-unified}} & \textbf{83.52} & \textbf{87.07} & 75.48 & - \\

\hline
\hline
\textbf{AAM~\cite{sukhbaatar2019adaptive}} & \textbf{79.27} & 83.49 & 76.45 & 79.34 \\
\textbf{AMOM~\cite{xiao2023amom}} & \textbf{78.13} & 82.98 & \textbf{77.49} & \textbf{82.09} \\

\hline
\textbf{ACTM-ATE} & \textbf{80.34} & 82.91 & \textbf{77.09} & \textbf{81.04} \\
\hline
\end{tabular}
\end{table}

\begin{table}[h]
\small
\centering
\caption{Performance of adaptive masking strategies against baseline models on ASC task. "\textbf{**}" indicates results reported in the paper as mean values.}
%For ACTM model, Gradient-based Mean is applied. "\textbf{-}" and "\textbf{*}" denotes the same from Table ~\ref{tab:ate_datasets} and
\label{tab:asc_datasets}
\begin{tabular}{|l|c|c|c|c|c|c|c|c|}
\hline
\multirow{2}{*}{\textbf{Model}} & \multicolumn{2}{c|}{Laptop14} & \multicolumn{2}{c|}{Restaurant14} & \multicolumn{2}{c|}{Restaurant15} & \multicolumn{2}{c|}{Restaurant16} \\ \cline{2-9} 
                       & Acc. & F1 & Acc. & F1 & Acc. & F1 & Acc. & F1 \\ \hline

\textbf{LCF-BERT-CDW~\cite{zeng2019lcf}}   & 82.45 & 79.59 & 87.14 & 81.74 & - & - & - & - \\
\textbf{LCF-BERT-CDM~\cite{zeng2019lcf}}   & 82.29 & 79.28 & 86.52 & 80.40 & - & - & - & - \\

\textbf{LCFS-ASC-CDW~\cite{phan-ogunbona-2020-modelling}}  & 80.52 & 77.13 & 86.71 & 80.31 & 89.03* & 73.31* & 92.25* & 76.46* \\
\textbf{LCFS-ASC-CDM~\cite{phan-ogunbona-2020-modelling}}  & 80.34 & 76.45 & 86.13 & 80.10 & 88.61* & 69.32* & 91.84* & 70.67* \\

\textbf{AM Weight-BERT**~\cite{feng2022unrestricted}} & 79.78 & 76.20 & 85.66 & 79.92 & - & - & - & - \\
\textbf{AM Word-BERT**~\cite{feng2022unrestricted}} & 79.87 & 76.26 & 85.57 & 79.02 & - & - & - & - \\
\textbf{Unified Generative~\cite{yan-etal-2021-unified}} & - & 76.76 & - & 75.56 & - & 73.91 & - & - \\
\textbf{MGGCN-BERT~\cite{xiao2022multi}}    & 79.57 & 76.30 & 83.21 & 75.38 & 82.90 & 69.27 & 89.66 & 73.99 \\
\textbf{AMA-GLCF**~\cite{lin2023adaptive}}    & - & 76.78 & - & 79.33 & - & - & - & 77.08 \\
\hline
\hline
\textbf{AAM~\cite{sukhbaatar2019adaptive}} & 82.51 & \textbf{79.61} & 84.92 & 81.71 & 89.34 & \textbf{75.13} & 90.86 & 77.18 \\
\textbf{AMOM~\cite{xiao2023amom}} & 81.61 & 79.09 & 85.95 & 80.10 & \textbf{91.50} & \textbf{74.14} & 92.17 & 76.73 \\

\hline
\textbf{ACTM-ASC} & \textbf{83.65} & 76.29 & \textbf{91.05} & \textbf{82.01} & \textbf{90.54} & \textbf{74.07} & \textbf{93.49} & \textbf{78.19} \\
\hline
\end{tabular}
\end{table}

%\subsection{Results}
\noindent \textbf{\textit{Discussion on ATE and ASC results}:} We present the performance of the proposed adaptive contextual masking strategies for the ATE task in Table~\ref{tab:ate_datasets}. We note that both the \emph{AMOM} and \emph{ACTM} versions of the adaptive contextual masking strategies outperform the baseline methods in three datasets. We also note that the \emph{AAM} version shows competitive but not leading performance, indicating its partial effectiveness in contextual understanding for ATE tasks. We also emphasize that the proposed ACTM strategy is competitive in two datasets due to its capability to understand nuanced contextual interpretation by setting adaptive thresholds for the masking function. Similarly, we give the detailed performance of the proposed adaptive strategies for ASC task in table~\ref{tab:asc_datasets}. Unlike the ATE task, we note a significant performance gain with both Accuracy (\%) and F1 in the adaptive contextual masking strategies in all datasets. Most importantly, the proposed ACTM strategy outperforms the other two adaptive strategies in three datasets and gives a competing performance in the Restaurant 15 dataset. Overall, we signify that our idea of adaptively masking based on local text tokens on top of the global contextual representations can achieve better results in standalone ABSA tasks considered in this work. While ACTM strategy leads the performance in ASC and is competitive in ATE, the AMOM strategy is promising with ATE tasks.

\definecolor{maxblue}{rgb}{0,0,1}

% Attention scores for reference (from the provided data)
\newcommand{\maxscore}{0.0683} % The highest attention score (for 'but')

\begin{table}[h]
\centering % Center the table on the page
\caption{F1 of ATE and ASC with multiple \textbf{Aggregator} operations. "\textbf{Gradient-based}" indicates that {\textbf{$\alpha, \gamma$}} are learnable parameters. Otherwise $\alpha=\gamma=1$}
\label{tab:combined_ablation}
\resizebox{\textwidth}{!}{% Resize table to fit within the page width
\small % Decreasing the font size
\begin{tabular}{|l|c|c|c|c|c|c|c|c|}
\hline
& \multicolumn{2}{c|}{\textbf{Laptop14}} & \multicolumn{2}{c|}{\textbf{Restaurant14}} & \multicolumn{2}{c|}{\textbf{Restaurant15}} & \multicolumn{2}{c|}{\textbf{Restaurant16}} \\ \cline{2-9} 
\multirow{-2}{*}{\textbf{Aggregate Functions}} & \multicolumn{1}{c|}{\textbf{ATE}} & \multicolumn{1}{c|}{\textbf{ASC}} & \multicolumn{1}{c|}{\textbf{ATE}} & \multicolumn{1}{c|}{\textbf{ASC}} & \multicolumn{1}{c|}{\textbf{ATE}} & \multicolumn{1}{c|}{\textbf{ASC}} & \multicolumn{1}{c|}{\textbf{ATE}} & \multicolumn{1}{c|}{\textbf{ASC}} \\

\hline
\textbf{Mean}        &  75.74   & 71.53          &  78.43  & 78.90          & 76.19 &  74.19& 80.10 &  74.39\\
\textbf{Median}        &  78.18   & 72.67          &  79.94  & 79.59          & 76.23 &  74.23& 79.01 &  71.14\\
\textbf{SD}            &   80.29 & 74.37          &  78.15  & 80.10          & 76.96 &  77.12 & \textbf{81.92} &  \textbf{78.26}\\
\textbf{Gradient-based Mean} &  \textbf{80.34}  & \textbf{76.29} & \textbf{82.91}   & 82.01          & \textbf{77.09} &  74.07& \textbf{81.04} &  \textbf{78.19}\\
\textbf{Gradient-based Median} & 76.05 & 73.72          &   80.17 & \textbf{83.06} & 75.73 &  \textbf{77.86}& 78.15 &  72.24\\
\textbf{Gradient-based SD}   &  77.13  & 73.21          &  75.13  & 78.15          & 75.86 &  75.38 & 76.91 &  72.78\\         
\hline
\end{tabular}
}
\end{table}

%\subsection{Discussion on Ablation Study}
\noindent \textbf{\textit{Discussion on Ablation Study}:} Since the ACTM has significant performance in both ATE and ASC tasks, we compare several aspects of the proposed ACTM strategy as the ablation study. In this study, we explore different \textbf{Aggregate} operators, like \emph{Mean}, \emph{Media}, and \emph{Standard Deviation (SD)}, along with variations that use only a constant weight $\alpha = \gamma = 1$. We present our results of our ablation study in Table~\ref{tab:combined_ablation}. Alhough we note that our default setting of \emph{Mean} aggregator with learnable $\alpha$ and $\gamma$ performs overall good in most of the cases, we find some exceptions. We note that a simple \emph{SD} aggregator with constant weight is able to match our default gradient based Mean aggregator for the latest Restaurant'16 dataset in both ATE and ASC tasks. Similarly, the gradient based Median aggregator is also able to give better performance in two ASC tasks. 

\noindent \textbf{\textit{Discussion on Qualitative Analysis}:} We qualitatively analyze the performance of ACTM strategy for the ATE task to evaluate the efficacy of the proposed method in identifying optimal aspects from the text. Table~\ref{subtab:attention_scores} demonstrates an example token masking by the proposed ACTM strategy and Table~\ref{subtab:threshold_comparison} compares ATE task using adaptive threshold and fixed threshold. It is evident from these tables that the proposed ACTM strategy can assist the ATE task to capture nuanced aspect terms using adaptive contextual masking technique.

\begin{table}[h]
    \caption{Qualitative data for Aspect Term Extraction}
    \label{tab:qualitative_ate}
    \centering

    \begin{subtable}{.35\textwidth}
        \caption{Tokens with Attention Scores and Masking Status}
        \label{subtab:attention_scores}
        \centering
        \small % Adjust the font size as necessary
        \setlength{\tabcolsep}{3pt} % Adjust the padding between table columns
        \renewcommand{\arraystretch}{1} % Adjust the padding between table rows
        \begin{tabular}{|m{1.3cm}|m{1.1cm}|m{1.2cm}|}
            \hline
            \textbf{Token} & \textbf{Attn. Score} &  \textbf{Masked} \\
            \hline
            \cellcolor{orange!46} the & 0.0460 & \cellcolor{red!20} Yes \\
            \hline
            \cellcolor{orange!100} steak & 0.1082 & \cellcolor{green!20} No \\
            \hline
            \cellcolor{orange!52} was & 0.0561 & \cellcolor{red!20} Yes \\
            \hline
            \cellcolor{orange!80} incredibly & 0.0867 & \cellcolor{green!20} No \\
            \hline
            \cellcolor{orange!72} tender & 0.0775 & \cellcolor{green!20} No \\
            \hline
            \cellcolor{orange!30} and & 0.0323 & \cellcolor{red!20} Yes \\
            \hline
            \cellcolor{orange!25} flavor & 0.0265 & \cellcolor{red!20} Yes \\
            \hline
            \cellcolor{orange!30} ful & 0.0319 & \cellcolor{red!20} Yes \\
            \hline
            \cellcolor{orange!26} , & 0.0275 & \cellcolor{red!20} Yes \\
            \hline
            \cellcolor{orange!90} but & 0.0977 & \cellcolor{green!20} No \\
            \hline
            \cellcolor{orange!74} service & 0.0794 & \cellcolor{green!20} No \\
            \hline
            \cellcolor{orange!38} quite & 0.0413 & \cellcolor{red!20} Yes \\
            \hline
            \cellcolor{orange!60} slow & 0.0648 & \cellcolor{green!20} No \\
            \hline
            \cellcolor{orange!46} . & 0.0493 & \cellcolor{red!20} Yes \\
            \hline
            \multicolumn{3}{|m{3cm}|}{\textbf{Total:} 0.8250} \\
            \hline
            \multicolumn{3}{|m{3cm}|}{\textbf{Mean:} 0.0590} \\
            \hline
        \end{tabular}
    \end{subtable}%
    \hfill
    \begin{subtable}{.60\textwidth}
        \caption{Comparison of ATE using Predefined vs. Dynamic Threshold for sample texts.}
        \label{subtab:threshold_comparison}
        \centering
        \small % Adjust the font size as necessary
        \setlength{\tabcolsep}{3pt} % Adjust the padding between table columns
        \renewcommand{\arraystretch}{1} % Adjust the padding between table rows
        \begin{tabular}{|m{3cm}|m{1.7cm}|m{1.7cm}|}
            \hline
            \textbf{Review Instances} & \textbf{ATE w/ fixed threshold} & \textbf{ATE w/ ACTM} \\
            \hline
            After numerous attempts of trying (including setting the \textit{\textbf{clock in BIOS setup}} directly), I gave up (I am a techie). & clock & clock in BIOS setup \\
            \hline
            After really enjoying ourselves at the \textit{\textbf{bar}} we sat down at a \textit{\textbf{table}} and had \textit{\textbf{dinner}}. & bar, dinner & bar, table, dinner \\
            \hline
            Did not enjoy the new \textbf{\textit{Windows 8}} and \textbf{\textit{touchscreen} functions}. & windows, touchscreen & windows 8, touchscreen functions\\
            \hline
        \end{tabular}
    \end{subtable}
\end{table}

\section{Conclusion}
In this work, we explored Aspect-based Sentiment Analysis (ABSA) with a focus on standalone tasks such as Aspect Term Extraction (ATE) and Aspect Sentiment Classification (ASC) using three different adaptive masking strategies. We introduced one of those strategies named Adaptive Contextual Threshold Masking (ACTM) while utilizing two other adaptive masking techniques for ABSA. We depicted with our experiments on benchmark datasets that adaptive masking can increase the chance of precise ATE and ASC tasks. Particularly, our ACTM strategy demonstrated significant effectiveness over other approaches with its adaptive contextual threshold module. For future research, we recommend investigating adaptive masking for both standalone and compound ABSA tasks which benefits many applications. Another open venue for improvement is to explore the adaptive masking for multi-modal ABSA tasks.
%Our work makes a substantial contribution to the ABSA field and sets the stage for further advancements in this area.

\bibliographystyle{unsrtnat}
\bibliography{references}  %%% Uncomment this line and comment out the ``thebibliography'' section below to use the external .bib file (using bibtex) .





% \end{thebibliography}

\end{document}